\def\year{2018}\relax
\documentclass[letterpaper]{article} %DO NOT CHANGE THIS
\usepackage{aaai18}  %Required
\usepackage{times}  %Required
\usepackage{helvet}  %Required
\usepackage{courier}  %Required
\usepackage{url}  %Required
\usepackage{graphicx}  %Required

\usepackage{bigstrut}
\usepackage{booktabs}

\begin{document}
\title{Unsupervised Part-based Weighting Aggregation of Deep Convolutional Features for Image Retrieval}
\author{Jian Xu,
Cunzhao Shi,
Chengzuo Qi,
Chunheng Wang\thanks{Corresponding author},
Baihua Xiao\\
State Key Laboratory of Management and Control for Complex Systems,\\
Institute of Automation, Chinese Academy of Sciences(CASIA)\\
%SKL-MCCS, CASIA\\
University of Chinese Academy of Sciences\\
\{xujian2015, cunzhao.shi, qichengzuo2013, chunheng.wang, baihua.xiao\}@ia.ac.cn
%\thanks{ (Corresponding author: Chunheng Wang)}% <-this % stops a space
}
\maketitle
\begin{abstract}

In this paper, we propose a simple but effective semantic part-based weighting aggregation (PWA) for image retrieval. The proposed PWA utilizes the discriminative filters of deep convolutional layers as part detectors. Moreover, we propose the effective unsupervised strategy to select some part detectors to generate the ``probabilistic proposals'', which highlight certain discriminative parts of objects and suppress the noise of background. The final global PWA representation could then be acquired by aggregating the regional representations weighted by the selected "probabilistic proposals" corresponding to various semantic content. We conduct comprehensive experiments on four standard datasets and show that our unsupervised PWA outperforms the state-of-the-art unsupervised and supervised aggregation methods. Code is available at \url{https://github.com/XJhaoren/PWA}.

\end{abstract}

\section{Introduction}%偏向介绍自己的动机，需要修改，体现别人的优势和不足（别人的方法可以一句话评述）
%%%传统聚合方法
%cnn不训练聚合方法
%cnn训练聚合方法
%本文的贡献
Over the past decades, image retrieval has received sustained attention.
The general retrieval framework~\cite{retrieval_survey} consists of some pivotal modules, i.e., image representation~\cite{rvd,rmac}, database indexing~\cite{IMI}, image scoring~\cite{C_R_are_one,DSM} and search reranking~\cite{three_thing}.
Image representations derived by aggregating features such as Scale-Invariant Feature Transform (SIFT)~\cite{sift} and Convolutional Neural Network (CNN)~\cite{feature_map} are shown to be effective for image retrieval~\cite{bow,vlad,fv_cvpr,tri_embed,faemb,rvd,nc,mr,spoc,rmac,crow,interactive,SCDA}.

%The recent advances are achieved by the aggregation of CNN features rather than SIFT features.
Recently, the performance of CNN-based features aggregation methods~\cite{nc,mr,spoc,rmac,crow} rapidly outperforms that of SIFT-based features aggregation methods~\cite{bow,vlad,fv_cvpr,fv_eccv,tri_embed,faemb,rvd}.
Some methods~\cite{off_the_shelf,msop,nc} generate the global representation based on fully connected layer features for image retrieval.
After that, convolutional features are aggregated to obtain the global representation~\cite{mr,spoc,rmac,crow,interactive} and achieve better performance.
%These methods use the pre-trained CNN which is not very suitable for image retrieval task.
%The pre-trained CNN does not perform very well for image retrieval task,
%therefore many recent works~\cite{netvlad,fine_tune_1,fine_tune_2} re-train the CNNs for image retrieval task  by collected landmark buildings datasets.
Many recent methods~\cite{netvlad,fine_tune_1,fine_tune_2,fine_tune_3} re-train the image representations end-to-end for image retrieval task  by collected landmark buildings datasets.
The fine-tuning process significantly improves the adaptation ability for the  specific  task.
However, these methods~\cite{netvlad,fine_tune_1,fine_tune_2,fine_tune_3}  need to collect the labeled training datasets and the performance heavily relies  on the collected datasets.
%NetVLAD~\cite{netvlad} proposes a new generalized VLAD~\cite{vlad} layer and re-trains the model  via the weakly supervised ranking loss, of which the input is the feature map of convolutional layer and the output is a global representation.
%~\cite{fine_tune_1,fine_tune_2} fine-tune deep CNN features for image retrieval which use pairwise and ranking loss respectively. After fine-tuning, they aggregate the CNN features by R-MAC~\cite{rmac} based method.
%However, the performance of these methods~\cite{netvlad,fine_tune_1,fine_tune_2} depends on the collected labeled training datasets.
The discrepant retrieval objects need different training datasets, for example, the fine-tuned model based on landmarks  is not suitable for logo retrieval.

\begin{figure}
  \centering
  \includegraphics[width=2.5 in]{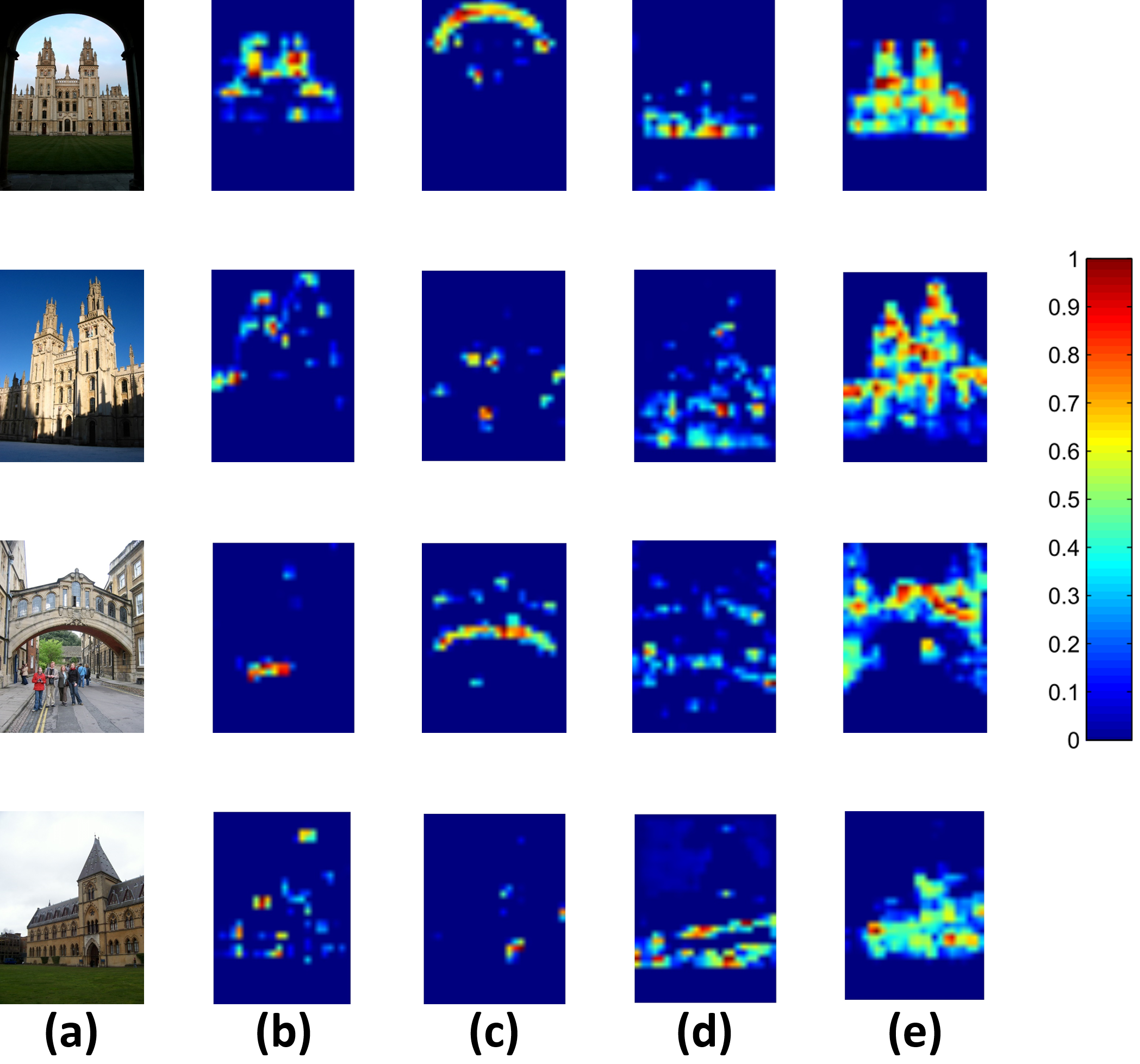}\\
  \caption{Visualization of ``probabilistic proposals''. (a) Some images in Oxford5K~\cite{oxford}. (b)-(e) The various channels of feature maps  in $pool5$ layer from pre-trained VGG16~\cite{VGG}. Each channel of feature maps is activated (warm) by different parts or patterns of objects and some discriminative channels can work as ``probabilistic proposals''.
  %For example, the (b) 220th feature map  is most activated by the sharp shape; the (c) 478th  feature map  is most activated by the arc shape; the (d) 483th feature map  is most activated by the bottom of buildings; the (e) 360th feature map  is most activated by the body of buildings.
  }
  \label{feature_map}
\end{figure}

Previous aggregation methods ignore the discriminative information from the object parts.
The part-based information is utilized for fine-gained categorization~\cite{filter_response,two_level_attention,unsupervised_part_model,part_based_representation,part_selection_spatial} and the part-based representation provides the state-of-the-art performance.
Zhang et al.~\cite{filter_response} pick some distinctive filters which respond to specific patterns significantly and consistently to learn a set of  part detectors. Then, they conditionally  encode the deep filter responses into the final representation based on Fisher vector~\cite{fv_cvpr}.
%The work ~\cite{two_level_attention} uses clustered mid-layer filters to detect parts from region proposals and builds the part-based classifier.
%%The result of~\cite{two_level_attention} shows that the part-based representation is discriminative.
%Simon et al.~\cite{unsupervised_part_model} interpret the outputs of convolutional layers as detection scores of multiple object part detectors, and select some important outputs from   layers of CNN as parts.
In recent work~\cite{part_based_representation}, the part-based image representation is generated by aggregating selected parts on several different scales.
The recent work~\cite{part_selection_spatial} applies spatial constraints to select part proposals which are generated by selective search~\cite{selective_search}.
Different with these methods, the selected parts proposals (``probabilistic proposals'') in our algorithm are not constrained to rectangular box but erose shape.

Some recent works~\cite{sppnet,filter_response,visualize} analyze the meaning of feature maps of CNN.
Zeiler et al.~\cite{visualize} show that some input patterns stimulate the special channels of feature maps of the latter convolutional layers.
He et al. visualize the feature maps  generated by some filters of the $conv_{5}$ layer from SPP-net~\cite{sppnet} and show that the filters of deep convolutional layers are activated by specific semantic content and some distinctive filters can work as  part detectors.
The various channels of convolutional feature maps can represent the  pixel-level label mask of different categories in Fully Convolutional Network (FCN)~\cite{fcn}.
Instance-aware semantic segmentation~\cite{instance2016,instance2017} employs the different channels of shared convolutional layers to detect and segment the various object instance jointly.
Mask R-CNN~\cite{mask_rcnn} demonstrates that the erose proposals perform better than the  rectangular regions on object detection task.
Inspired by above works, we employ some selected discriminative filters of deep convolutional layers as the part detectors to generate erose ``probabilistic proposals'', which correspond to  fixed semantic content implicitly.

In this paper, we define the special channel of normalized feature maps as  ``probabilistic proposal''.
The ``probabilistic proposal'' encodes the spatial layout of input object's parts corresponding to various semantic content, and represents the  probability of pixels belonging to fixed semantic.
To further understand  the meanings and characteristics  of the ``probabilistic proposals'', we visualize some images and  corresponding typical ``probabilistic proposals'' in Fig.~\ref{feature_map}.
We select some images in Oxford5K~\cite{oxford} as shown in Fig.~\ref{feature_map} (a).
In Fig.~\ref{feature_map} (b)-(e), we visualize some discriminative channels of feature maps which work as the ``probabilistic proposals''  for the selected images.
Each channel of feature maps is activated (warm) by special parts or patterns corresponding to fixed semantic content and the background is suppressed (cold).
For example, the 220th feature map (Fig.~\ref{feature_map} (b)) of $pool5$ layers from VGG16~\cite{VGG} is most activated by the sharp shape; the 478th feature map  (Fig.~\ref{feature_map} (c)) is most activated by the arc shape; the 483th feature map  (Fig.~\ref{feature_map} (d)) is most activated by the bottom of buildings; the 360th feature map (Fig.~\ref{feature_map} (e)) is most activated by the body of buildings.
We can see that different filters of deep convolutional layers are sensitive to different shapes or semantic, and they highlight different parts and patterns of objects.
Some special parts of object are discriminative, for example, the 220th feature maps highlight the spire of buildings.
Therefore, filters of deep convolutional layers can work as part detectors to pick special patterns corresponding to fixed semantic content.
We select the discriminative filters of deep convolutional layers as the part detectors to generate erose ``probabilistic proposals, which are related to different semantic content.
%The previous CNN-based aggregation methods ignore the part-based semantic information of feature maps. In this paper, we propose a succinct unsupervised PWA method considering part-based semantic information.

%However, the previous aggregation methods~\cite{off_the_shelf,msop,nc,mr,spoc,rmac,crow,netvlad,fine_tune_1,fine_tune_2,fine_tune_3} ignore the information of the object parts.
Inspired by the characteristics of feature maps, in this paper we propose a novel and simple way of creating powerful image representation via part-based aggregation.
%The proposed method uses the normalized values of special channels of feature maps as ``probabilistic proposals'' generated by selected part detectors.
Our unsupervised part-based weighting aggregation (PWA) method significantly outperforms the state-of-the-art unsupervised aggregation methods~\cite{mr,spoc,rmac,crow} and supervised methods~\cite{fine_tune_1,fine_tune_2,fine_tune_3} on four standard retrieval datasets.

The main contributions of this paper can be summarized as follows:

\subsubsection{``Probabilistic proposal''}
We select some discriminative part detectors by succinct unsupervised strategy to generate the ``probabilistic proposals'' corresponding to special semantic content.
Different with previous methods, the selected ``probabilistic proposals'' are not constrained to rectangular box and represent the confidence degree of fixed semantic.
To the best of our knowledge, this paper is the first work to select the erose ``probabilistic proposals'' for image retrieval, and the selected ``probabilistic proposals'' corresponding to special semantic content are tactfully employed to generate  high-dimensional  representation  which contains discriminative semantic information.

\begin{figure*}
  \centering
  \includegraphics[width=6 in]{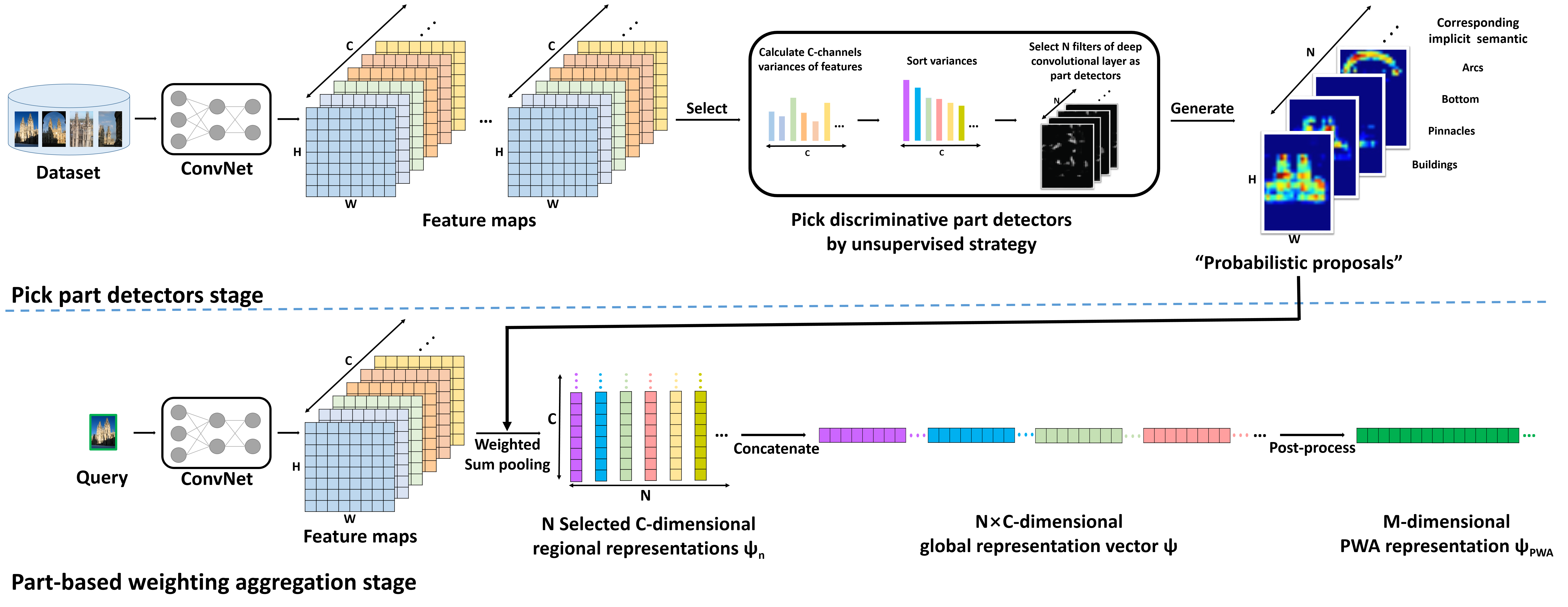}\\
  \caption{ Flow chart of our part-based weighting aggregation (PWA) method.
  We pick the discriminative part detectors  to generate the ``probabilistic proposals" by the unsupervised strategy in the first off-line stage.
  Each ``probabilistic proposal"  corresponds to fixed semantic content implicitly, such as pinnacles, arcs and bottom of buildings.
  In the aggregation stage, we employ the  selected N ``probabilistic proposals"  to weight and aggregate the feature maps as C-dimensional regional representations, and concatenate N regional representations as the final global PWA representation.
  }
  \label{chart_flow}
\end{figure*}

\subsubsection{Part-based weighting aggregation}
We aggregate the convolutional features weighted by selected ``probabilistic proposals'' and concatenate the regional representations as global PWA representation.
Because selected ``probabilistic proposals'' corresponds to fixed semantic but not fixed position, the selected regional representations can be concatenated as the global PWA representation.
Concatenation as the global representation preserves more discrimination than summing regional representations.

\section{Aggregation based on ``probabilistic proposals"}
The diagram of the proposed method is shown in Fig.~\ref{4}.
Based on the dataset, we pick the discriminative part detectors  to generate the ``probabilistic proposals" by the unsupervised strategy in the off-line stage.
Each ``probabilistic proposal" corresponds to fixed semantic content implicitly, such as pinnacles, arcs and bottom of buildings.
In the aggregation stage, we employ the  selected N ``probabilistic proposals"  to weight and aggregate the feature maps as C-dimensional regional representations.
Finally, we concatenate N regional representations corresponding to special sematic content as the final global PWA representation.

In this section, we analyse the characteristics of the filters of deep convolutional layers which can be interpreted as part detectors.
We propose the unsupervised strategy to select discriminative part detectors to generate  ``probabilistic proposals".
Based on the selected ``probabilistic proposals"  corresponding to special semantic content, we propose a novel and effective PWA aggregation method for image retrieval.

We extract  features $f$ from deep convolutional layers  by passing an image $I$ through a pre-trained  or fine-tuned deep network, which consist of $C$ channels feature maps each with height $H$ and width $W$. Finally, the input image $I$ is represented by the aggregated $N\times C$-dimensional vector that are weighted by the  $N$ selected part detectors.

%分析部件，即思想来源，该方法为什么有效

\subsection{``Probabilistic proposals"}

\subsubsection{Selection of part detectors}
%如何选择较好的feature map（鉴别性较强） 画图显示某些部件很有鉴别性，有些没有（用方差最大和最小的比较）
Because the responses  with large variances are significantly different among the various objects, the channels of feature maps with large variances are more discriminative. Therefore, we select part detectors according to variances based on dataset.

We first calculate the C-channels variances $V=\{v_{1}, v_{2}, ..., v_{c}, ..., v_{C}\}$  of the $C$-dimensional vectors $g_{i}$ ($i=1,2,...,D$) computed by sum pooling the  $C\times W\times H$-dimensional deep convolutional features $f_{i}$ of image $i$.
%公式化详细说明明怎么计算方差
\begin{equation}\label{0}
V = \frac{1}{D}\sum\limits_{i = 1}^D {({g_{_i}}}  - \bar g{)^2}
\end{equation}
where $D$ is the  number of database images. $\overline{g}=\frac{1}{D}\sum\limits_{i = 1}^D {g_{i}}$ is the average vector of feature vectors $g_{i}$ ($i=1,2,...,D$).
\begin{equation}\label{0}
{g_{i}} = \sum\limits_{x = 1}^W {\sum\limits_{y = 1}^H {f_{i}(x,y)} }
\end{equation}

Then we sort the variances $\{v_{1}, v_{2}, ..., v_{C}\}$ of C channels.
We select the discriminative deep convolutional layers filters corresponding to large variances as the part detectors.
We also observe the filters with large variances to be more discriminative by the following experiment.
We performed retrieval by PWA but we select  (1) 30\% random part detectors (2) 30\% part detectors with the largest variance. The mAP score for the Oxford5k dataset~\cite{oxford} for (1) is only 0.775$\pm$0.006, which is much small than mAP for (2), 0.790. This verifies that feature maps with large variances are much more discriminative than random feature maps. Moreover, our simple unsupervised selection method not only boosts the performance but also reduces the computational complexity of PWA representation.

\begin{figure*}
  \centering
  \includegraphics[width=7 in]{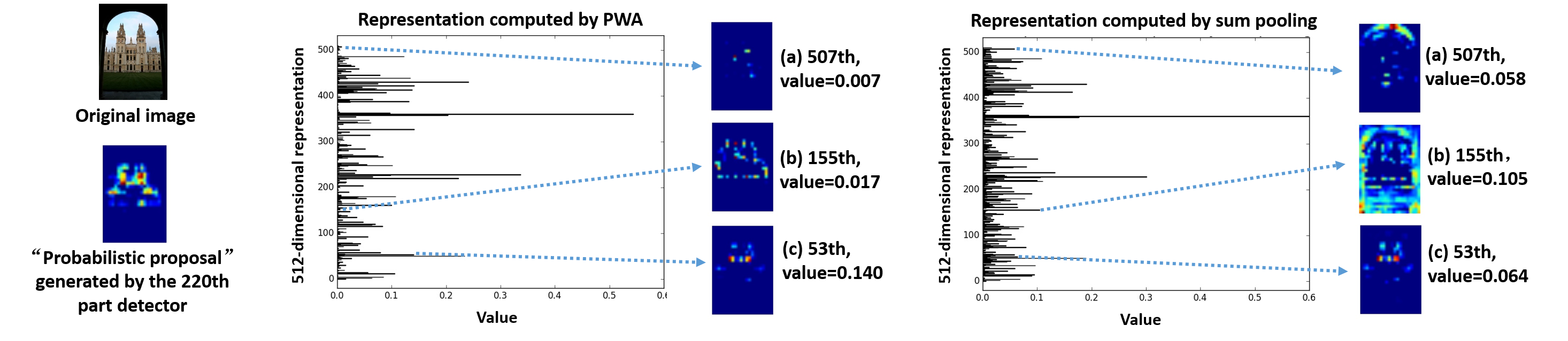}\\
  \caption{The comparison  of the 512-dimensional representations computed by PWA and sum pooling. Weighted by the selected ``probabilistic proposal'',   values of the feature map's channels activated by background (such as (a) 507th and (b) 155th) are reduced. However,  values of the representation  corresponding to similar patterns to the selected ``probabilistic proposal'' (such as (c) 53th) still keep large. The selected ``probabilistic proposal''  suppresses (cold) the  noise of background and highlights (warm)  the special semantic content.}
  \label{sumpooling_pwa}
\end{figure*}

\subsubsection{Effects of ``probabilistic proposals"}
%可视化加权后的feature map，证明为什么部件加权有效（使有区分性的部件更为突出，抑制噪声的作用）,插图比较抑制与增强
%In Fig.~\ref{weighted_feature_maps}, we visualize the feature maps which are weighted by the ``probabilistic proposal'' with the largest variance, which is the 360th  feature map of $pool5$ layer from VGG16~\cite{VGG}. We compare the original feature maps  (in Fig.~\ref{weighted_feature_maps} (c)) with the feature maps weighted by part detectors (in Fig.~\ref{weighted_feature_maps} (d)). The results show that the pivotal  parts of retrieval objects are highlighted (red) and the noise of CNN features is suppressed (blue) by the discriminative ``probabilistic proposals''.
%Therefore, The discriminative part detectors activate the corresponding patterns and effectively  suppress noise.

%分析加权后512*512维特征的分布（实验是否每512维不l2归一化更好）,抑制噪声（背景）的干扰，可以用all_souls_000002第510 张feature map 加权和不加权为例分析。抑制噪声干扰，增强前景的响应，使加权后的特征更有鉴别性。
The special channels of feature maps  generated by selected part detectors can work as the ``probabilistic proposals" corresponding to fixed semantic content.
To investigate the effects of ``probabilistic proposals" in detail, we compare the 512-dimensional representation computed by sum pooling with the representation weighted by the discriminative ``probabilistic proposals" in Fig.~\ref{sumpooling_pwa}.
As shown in Fig.~\ref{sumpooling_pwa}, the selected ``probabilistic proposal'' generated by 220th part detector suppresses the  noise of background and activates the sharp shape.
Weighted by the selected ``probabilistic proposal'', the values of feature maps that are activated by background (such as (a) 507th and (b) 155th) are smaller.
However, the values of the representation  corresponding to similar semantic content  to the selected ``probabilistic proposal'' (such as (c) 53th) still keep large.
As a result, the representations weighted by the discriminative ``probabilistic proposals'' are more discriminative and robust.
%加权后删除某些（方差较小）维度的特征不降低map，说明加权后这些噪声被抑制。
%We also conduct an additional experiment to prove that the noise of background can be suppressed by part detectors.
%We select the $N$ discriminative part detectors  and only use $N$ channels of feature maps responding to the selected part detectors to generate the (1) $N\times N$-dimensional  representation rather than (2) $N\times C$. The mAP of (1) is nearly same as (2). The result shows that the channels which are not selected are dramatically  suppressed  by part detectors and  almost have no effect on discrimination.

%很多部件重复（feature map表示近似的特征，可以选择聚成包，选择部件检测器）
Overall, the discriminative filters of latter convolutional layers are interpreted as part detectors to generate the ``probabilistic proposals''.
The selected ``probabilistic proposals'' suppress the  noise of background and highlight the discriminative  parts and patterns of objects.
We make use of the selected ``probabilistic proposals'' to weight the activations of convolutional layers and generate the regional representations.
%视作proposal +与r-mac和crow不同
Because each filter of deep convolutional layers activates special pattern, the various selected part detectors can be employed to generate the erose proposals  corresponding to special semantic content.
Each proposal corresponds to a fixed semantic pattern  implicitly.
The erose ``probabilistic proposals'' maintain the explicit $W\times H$ object spatial layout which can be addressed naturally by the pixel-to-pixel correspondence provided by convolutions.
%与rmac比proposal是对应固定语义的，而不是位置
Different with R-MAC~\cite{rmac}, the ``probabilistic proposals'' corresponds to fixed semantic content rather than fixed position.
Our ``probabilistic proposals'' are not constrained to box and represent the probability of pixels belong to fixed semantic content.
Although the ``probabilistic proposals''   corresponding to the part detectors selected by unsupervised strategy do not explicitly describe the semantic, they implicitly represent discriminative semantic content, such as pinnacles, arcs and bottom of buildings.
Therefore we can concatenate the selected regional representations weighted by  special semantic ``probabilistic proposals'' as the final global representation.
%Concatenation preserve more discriminative information than summing regional representations in R-MAC~\cite{rmac}.
%与crow比不是使用所有通道加和，而是独立使用挑选的part detector 去提取对应固定模式的proposal，并且将对应不同proposal 的representation串起来
%与interactive比较
CroW~\cite{crow}, InterActive~\cite{interactive} and PWA can be interpreted as spatial-weighted representations.
InterActive~\cite{interactive} is much more generalized in the aspect of spatial-weighted, which integrates high-level visual context with low-level neuron responses by back-propagation.
Compared to CroW~\cite{crow} and InterActive~\cite{interactive} that sum the spatial-weighted representations, we independently employ the selected part detectors to extract the regional representations corresponding to special semantic content and concatenate them as final PWA representation.
The concatenation of regional representations preserves more discriminative information than summation in R-MAC~\cite{rmac}, CroW~\cite{crow} and InterActive~\cite{interactive}.
%

%具体实现
\subsection{PWA design}
In this section, we describe the  PWA  method in detail.
We aggregate the feature maps weighted by the  selected ``probabilistic proposals'' and concatenate the regional representations as global PWA representation.
We reduce the dimensionality of  high-dimensional PWA representation by unsupervised method (PCA) in post-processing.

\subsubsection{Weighted by selected ``probabilistic proposals''} The construction of the PWA representation starts with the weighted sum pooling of the $C\times W\times H$-dimensional deep convolutional features $f$ of image $I$ with height $H$ and width $W$:
\begin{equation}\label{1}
{\psi _{\rm{n}}}(I) = \sum\limits_{x = 1}^W {\sum\limits_{y = 1}^H {{w_n}(x,y)} f(x,y)}
\end{equation}

The coefficients $w_{n}$ are the normalized weights as follows, which depend on the activation values $v_{n}(x,y)$ in position $(x,y)$ of the selected ``probabilistic proposal'' generated by part detector $n$:
\begin{equation}\label{2}
{w_n}(x,y) = {\left( {\frac{{{v_n}(x,y)}}{{{{(\sum\limits_{x = 1}^W {\sum\limits_{y = 1}^H {{v_n}} } {{(x,y)}^\alpha })}^{1/\alpha }}}}} \right)^{1/\beta }}
\end{equation}
where $\alpha$ and $\beta$ are parameters of power normalization and power-scaling respectively.

\subsubsection{Concatenation} $N$ selected $C$-dimensional regional representations $\psi_{n}(I)$ are obtained from weighted sum pooling process. We get the global $N\times C$-dimensional representation vector $\psi(I)$ by concatenating selected  regional representations:
\begin{equation}\label{3}
\psi (I) = \left[ {{\psi _1},{\psi _2}, \cdots {\psi _N}} \right]
\end{equation}
where we  select  the N  part detectors  depending on the discrimination of them. The selection based on the values of the variances of different  $C$ channels of feature maps both provides  boost in performance  and  enhances the computation efficiency.

\subsubsection{Post-processing} We  perform  $l_{2}$-normalization, PCA compression and whitening on the global representation $\psi(I)$ subsequently and obtain the final M-dimensional representation $\psi_{PWA}(I)$ :
\begin{equation}\label{4}
{\psi _{PWA}}(I) = diag{({\sigma _1},{\sigma _2}, \cdots ,{\sigma _M})^{ - 1}}{V}\frac{{\psi (I)}}{{{{\left. {\left\| {\psi (I)} \right.} \right\|}_2}}}
\end{equation}
where $V$ is the $M\times N$ PCA-matrix, $M$ is the number of the retained dimensionality, and ${\sigma _1},{\sigma _2}, \cdots ,{\sigma _M}$ are the associated singular values.

\section{Experiments}
\subsection{Datasets}
We evaluate the performance of PWA and other aggregation algorithms on four standard datasets (Oxford5k, Paris6k, Oxford105k and Paris106k) for image retrieval.

Oxford5k~\cite{oxford} and Paris6k~\cite{paris}  datasets contain photographs collected from Flickr associated with Oxford and Paris landmarks respectively.
The performance is measured using mean average precision (mAP) over the 55 queries  annotated manually.
Oxford105k and Paris106k  contain the additional 10,000 distractor images from Flicker~\cite{oxford}.
%Paris dataset~\cite{paris} (Paris6K) contains 6412 photographs from Flickr  associated with Paris landmarks. The performance is measured using mean average precision (mAP) over the 55 queries that are manually annotated.
%Paris dataset+100K (Paris106K) contains the Paris dataset~\cite{paris} and additionally 100K distractor images from Flicker~\cite{oxford}.
%Holidays dataset~\cite{holidays}  is a set of images which mainly contains 1491 personal holidays photos corresponding to 500 groups each having the same scene or object.
%Query images are the sets of the first image of each group and the correct retrieval results are the other images of the same group.
%The performance is reported as mean average precision (mAP) over 500 queries.
%Similarly to~\cite{nc,spoc}, we manually fix images in wrong orientation by rotating them by $\pm$90 degrees.

\subsection{Implementation details}
We extract deep convolutional features using the pre-trained VGG16~\cite{VGG} and fine-tuned ResNet101 from the work~\cite{fine_tune_3}. In the experiments, Caffe~\cite{caffe} package for CNNs is used. For VGG16 model, we extract convolutional feature maps from the $pool5$ layer and the number of channels is C=512. For ResNet-101 model, we extract convolutional feature maps from the $res5c_{-}relu$ layer and the number of channels is C=2048.
Regarding image size, we keep the original size of the images except for the very large images which are resized to the half size.
The parameters for power normalization and power-scaling are set as $\alpha=2$ and $\beta=2$, throughout our experiments.

We evaluate the mean average precision (mAP) over the cropped query.
For fair comparison with the related retrieval methods, we learn the PCA and whitening parameters on Oxford5k when testing on Paris6k and vice versa. %and we use Oxford5k dataset for whitening on the Holidays.

%(except top 2 for Holidays dataset, because the number of images in the same category is two mostly)
%QE consistently improves the performance on all datasets, although it has a negligible cost.

\begin{table}[htbp]
  \centering
  %\setlength{\abovecaptionskip}{0pt}%
%  \setlength{\belowcaptionskip}{0pt}%
  %\vspace{4em}
  \caption{Performance of different number of selected part detectors (N).
  We aggregate the responses of convolutional layers  by all the C=512 part detectors  as the baseline.
  %We still achieve good performance by employing a small number of part detectors (e.g., N=25).
  %The best performance is achieved by selecting N=25 part detectors.
  Note, the final representation $\psi_{PWA}(I)$ is reduced  into 4096 dimensionality by PCA.}
    \vspace{0.5em}
    \begin{tabular}{ccc}
    \toprule[1.25pt]
     %\hline
          & \multicolumn{2}{c}{\textbf{Datasets}} \bigstrut\\
\cline{2-3}    \multicolumn{1}{c}{\textbf{N}} & \multicolumn{1}{c}{Oxford5k} & Paris6k \bigstrut\\

    \hline
    \multicolumn{1}{c}{512} & \multicolumn{1}{c}{78.5} & \multicolumn{1}{c}{85.4} \bigstrut[t]\\
    \multicolumn{1}{c}{450} & \multicolumn{1}{c}{78.7} & \multicolumn{1}{c}{85.7} \\
    \multicolumn{1}{c}{350} & \multicolumn{1}{c}{79.0} & \multicolumn{1}{c}{85.9} \\
    \multicolumn{1}{c}{250} & \multicolumn{1}{c}{78.7} & \multicolumn{1}{c}{86.0} \\
    \multicolumn{1}{c}{150} & \multicolumn{1}{c}{79.0} & \multicolumn{1}{c}{85.4} \\
    \multicolumn{1}{c}{50} & \multicolumn{1}{c}{78.2} & \multicolumn{1}{c}{86.1} \\
    \multicolumn{1}{c}{25} & \multicolumn{1}{c}{\textbf{79.1}} & \multicolumn{1}{c}{\textbf{86.1}} \\
    \multicolumn{1}{c}{10} & \multicolumn{1}{c}{77.7} & \multicolumn{1}{c}{83.8} \\
    %\hline
    \bottomrule[1.25pt]
    \end{tabular}%
  \label{select}%
\end{table}%
\subsection{Impact of the parameters}
%Our methods only have few parameters to evaluate.
The main parameters are the numbers  of the selected part detectors  and the dimensionality  of final representations $\psi_{PWA}(I)$.

\subsubsection{Select part detectors}
We employ the discriminative filters of deep convolutional layers as part detectors to generate ``probabilistic proposals''.
The discriminative part detectors are selected according to the variances of C channels of feature maps.
We also aggregate the responses of convolutional layers  based on all the C part detectors  as the baseline.
We show the results of selecting the first N part detectors with the largest variance in Table~\ref{select}.
In this experiment, the final representation $\psi_{PWA}(I)$ is reduced  into 4096 dimensionality by PCA.

The results show that our PWA representation is not heavily relied on the number of selected part detectors.
Selecting a small number of part detectors (e.g., N=25), we still achieve good performance.
The selection strategy boosts above 0.6\% mAP than  baseline and reduces the computational cost to  1$/$20 of the baseline.
The results demonstrate that our straightforward unsupervised selection strategy is effective.

\begin{table}[htbp]
  \centering
  \caption{Performance of varying dimensionality (M), into which the final representation is reduced.
  The representation is reduced by PCA and whitening.
  %The best performance is achieved at 4096 dimensionality.
  Note, we select 25 part detectors to aggregate the convolutional features.}
   %\scalebox{1.5}{
    \vspace{0.5em}
    \begin{tabular}{ccc}
    \toprule[1.25pt]
    %\hline
          & \multicolumn{2}{c}{\textbf{Datasets}} \bigstrut\\
\cline{2-3}    \multicolumn{1}{c}{\textbf{M}} & \multicolumn{1}{c}{Oxford5k} & \multicolumn{1}{c}{Paris6k} \bigstrut\\
    \hline
    \multicolumn{1}{c}{128} & \multicolumn{1}{c}{64.5} & \multicolumn{1}{c}{76.9} \bigstrut[t]\\
    \multicolumn{1}{c}{256} & \multicolumn{1}{c}{68.7} & \multicolumn{1}{c}{79.6} \\
    \multicolumn{1}{c}{512} & \multicolumn{1}{c}{72.0} & \multicolumn{1}{c}{82.3} \\
    \multicolumn{1}{c}{1024} & \multicolumn{1}{c}{75.3} & \multicolumn{1}{c}{84.2} \\
    \multicolumn{1}{c}{2048} & \multicolumn{1}{c}{78.2} & \multicolumn{1}{c}{85.4} \\
    \multicolumn{1}{c}{4096} & \multicolumn{1}{c}{\textbf{79.1}} & \multicolumn{1}{c}{\textbf{86.1}} \bigstrut[b]\\
    %\hline
    \bottomrule[1.25pt]
    \end{tabular}%
   %}
  \label{dim}%
\end{table}%

\begin{table*}[htbp]
  \centering
  \caption{Accuracy comparison with the state-of-the-art unsupervised methods. We compare our PWA+QE  with other methods followed by query expansion at the bottom of table. Part-based weighting aggregation (PWA) consistently outperforms the state-of-the-art unsupervised aggregation methods.
  %We do not perform QE on Holidays as it is not a standard practice.
  }
    \vspace{0.5em}
    \begin{tabular}{rrrrrr}
    \toprule[1.5pt]
          &       & \multicolumn{4}{c}{\textbf{Datasets}} \bigstrut[b]\\
\cline{3-6}    \multicolumn{1}{l}{\textbf{Method}} & \multicolumn{1}{c}{\textbf{Dimensionality}} & \multicolumn{1}{c}{Oxford5k} & \multicolumn{1}{c}{Paris6k} & \multicolumn{1}{c}{Oxford105k} & \multicolumn{1}{c}{Paris106k}  \bigstrut[t]\\
    \toprule[1pt]
    %\multicolumn{1}{l}{Tri-embedding~\cite{tri_embed}} & \multicolumn{1}{c}{1024} & \multicolumn{1}{c}{56.0 } & \multicolumn{1}{c}{---} & \multicolumn{1}{c}{50.2 } & \multicolumn{1}{c}{---}  \\
    \multicolumn{1}{l}{Tri-embedding~\cite{tri_embed}} & \multicolumn{1}{c}{8k} & \multicolumn{1}{c}{67.6 } & \multicolumn{1}{c}{---} & \multicolumn{1}{c}{61.1 } & \multicolumn{1}{c}{---}  \\
    %\multicolumn{1}{l}{FAemb~\cite{faemb}} & \multicolumn{1}{c}{8k} & \multicolumn{1}{c}{66.7 } & \multicolumn{1}{c}{---} & \multicolumn{1}{c}{---} & \multicolumn{1}{c}{---}  \\
    \multicolumn{1}{l}{FAemb~\cite{faemb}} & \multicolumn{1}{c}{16k} & \multicolumn{1}{c}{70.9 } & \multicolumn{1}{c}{---} & \multicolumn{1}{c}{---} & \multicolumn{1}{c}{---}  \\
  %  \multicolumn{1}{l}{RVD-W~\cite{rvd}} & \multicolumn{1}{c}{8k} & \multicolumn{1}{c}{66.8 } & \multicolumn{1}{c}{---} & \multicolumn{1}{c}{64.0 } & \multicolumn{1}{c}{---}  \\
    \multicolumn{1}{l}{RVD-W~\cite{rvd}} & \multicolumn{1}{c}{16k} & \multicolumn{1}{c}{68.9 } & \multicolumn{1}{c}{---} & \multicolumn{1}{c}{66.0 } & \multicolumn{1}{c}{---}  \\
    \multicolumn{1}{l}{Razavian et al.~\cite{mr}} & \multicolumn{1}{c}{512} & \multicolumn{1}{c}{46.2 } & \multicolumn{1}{c}{67.4 } & \multicolumn{1}{c}{--- } & \multicolumn{1}{c}{---}  \\
    \multicolumn{1}{l}{Neural Codes~\cite{nc}} & \multicolumn{1}{c}{512} & \multicolumn{1}{c}{43.5 } & \multicolumn{1}{c}{---} & \multicolumn{1}{c}{39.2 } & \multicolumn{1}{c}{---}  \\
    \multicolumn{1}{l}{SPoC~\cite{spoc}} & \multicolumn{1}{c}{256} & \multicolumn{1}{c}{53.1 } & \multicolumn{1}{c}{---} & \multicolumn{1}{c}{50.1 } & \multicolumn{1}{c}{---}  \\
    \multicolumn{1}{l}{InterActive~\cite{interactive}} & \multicolumn{1}{c}{512} & \multicolumn{1}{c}{65.6} & \multicolumn{1}{c}{79.2} & \multicolumn{1}{c}{---} & \multicolumn{1}{c}{---}  \\
    \multicolumn{1}{l}{R-MAC~\cite{rmac}} & \multicolumn{1}{c}{512} & \multicolumn{1}{c}{66.9 } & \multicolumn{1}{c}{83.0 } & \multicolumn{1}{c}{61.6 } & \multicolumn{1}{c}{75.7 }  \\
    \multicolumn{1}{l}{CroW~\cite{crow}} & \multicolumn{1}{c}{512} & \multicolumn{1}{c}{70.8 } & \multicolumn{1}{c}{79.7 } & \multicolumn{1}{c}{65.3 } & \multicolumn{1}{c}{72.2 }  \\
    \hline
    \multicolumn{1}{l}{Previous state-of-the-art} & \multicolumn{1}{c}{} & \multicolumn{1}{c}{70.8 } & \multicolumn{1}{c}{83.0 } & \multicolumn{1}{c}{65.3 } & \multicolumn{1}{c}{75.7 }  \\
    \hline
    \multicolumn{1}{l}{PWA} & \multicolumn{1}{c}{512} & \multicolumn{1}{c}{\textbf{72.0}} & \multicolumn{1}{c}{82.3} & \multicolumn{1}{c}{\textbf{66.2}} & \multicolumn{1}{c}{\textbf{75.8}}  \\
    \multicolumn{1}{l}{PWA} & \multicolumn{1}{c}{1024} & \multicolumn{1}{c}{75.3} & \multicolumn{1}{c}{84.2} & \multicolumn{1}{c}{69.3} & \multicolumn{1}{c}{78.2} \\
    \multicolumn{1}{l}{PWA} & \multicolumn{1}{c}{2048} & \multicolumn{1}{c}{78.2} & \multicolumn{1}{c}{85.4} & \multicolumn{1}{c}{71.1} & \multicolumn{1}{c}{79.7}  \\
    \multicolumn{1}{l}{PWA} & \multicolumn{1}{c}{4096} & \multicolumn{1}{c}{\textbf{79.1}} & \multicolumn{1}{c}{\textbf{86.1}} & \multicolumn{1}{c}{\textbf{73.6}} & \multicolumn{1}{c}{\textbf{80.4}}  \\
    \toprule[1pt]
   % \toprule[1pt]
    %\multicolumn{1}{l}{Chum et al.~\cite{qe_1chum}} & \multicolumn{1}{l}{} & \multicolumn{1}{l}{82.7} & \multicolumn{1}{l}{80.5} & \multicolumn{1}{l}{76.7} & \multicolumn{1}{l}{71} & \multicolumn{1}{l}{---} \\
%    \multicolumn{1}{l}{Danfeng et al.~\cite{qe_2danfeng}} & \multicolumn{1}{l}{} & \multicolumn{1}{l}{81.4} & \multicolumn{1}{l}{80.3} & \multicolumn{1}{l}{76.7} & \multicolumn{1}{l}{---} & \multicolumn{1}{l}{---} \\
%    \multicolumn{1}{l}{Mikulik et al.~\cite{qe_3mikulik}} & \multicolumn{1}{l}{} & \multicolumn{1}{l}{84.9} & \multicolumn{1}{l}{82.4} & \multicolumn{1}{l}{79.5} & \multicolumn{1}{l}{77.3} & \multicolumn{1}{l}{75.8} \\
%    \multicolumn{1}{l}{Shen et al.~\cite{qe_4shen}} & \multicolumn{1}{l}{} & \multicolumn{1}{l}{75.2} & \multicolumn{1}{l}{74.1} & \multicolumn{1}{l}{72.9} & \multicolumn{1}{l}{---} & \multicolumn{1}{l}{76.2} \\
%    \multicolumn{1}{l}{Tao et al.~\cite{qe_5tao}} & \multicolumn{1}{l}{} & \multicolumn{1}{l}{77.8} & \multicolumn{1}{l}{---} & \multicolumn{1}{l}{---} & \multicolumn{1}{l}{---} & \multicolumn{1}{l}{78.7 } \\
%    \multicolumn{1}{l}{Tolias et al.~\cite{qe_6tolias}} &       & \multicolumn{1}{l}{\textbf{86.9}} & \multicolumn{1}{l}{85.1} & \multicolumn{1}{l}{\textbf{85.3}} & \multicolumn{1}{l}{---} & \multicolumn{1}{l}{81.3} \\
%    \hline

    \multicolumn{1}{l}{CroW+QE~\cite{crow}} & \multicolumn{1}{c}{512} & \multicolumn{1}{c}{74.9} & \multicolumn{1}{c}{84.8} & \multicolumn{1}{c}{70.6} & \multicolumn{1}{c}{79.4} \bigstrut[b]\\
    \multicolumn{1}{l}{R-MAC+AML+QE~\cite{rmac}} & \multicolumn{1}{c}{512} & \multicolumn{1}{c}{77.3} & \multicolumn{1}{c}{86.5} & \multicolumn{1}{c}{73.2} & \multicolumn{1}{c}{79.8}  \\
    \multicolumn{1}{l}{DSM~\cite{DSM}} & \multicolumn{1}{c}{---} & \multicolumn{1}{c}{95.0} & \multicolumn{1}{c}{91.5} & \multicolumn{1}{c}{93.2} & \multicolumn{1}{c}{---}  \\
    \multicolumn{1}{l}{PWA+QE} & \multicolumn{1}{c}{512} & \multicolumn{1}{c}{{74.8}} & \multicolumn{1}{c}{{86.0}} & \multicolumn{1}{c}{{72.5}} & \multicolumn{1}{c}{{80.7}} \\
    \multicolumn{1}{l}{PWA+QE} & \multicolumn{1}{c}{1024} & \multicolumn{1}{c}{{77.9}} & \multicolumn{1}{c}{{87.8}} & \multicolumn{1}{c}{{76.7}} & \multicolumn{1}{c}{{82.8}} \\
    \multicolumn{1}{l}{PWA+QE} & \multicolumn{1}{c}{2048} & \multicolumn{1}{c}{{80.7}} & \multicolumn{1}{c}{{88.7}} & \multicolumn{1}{c}{{79.3}} & \multicolumn{1}{c}{{83.9}}\\
    \multicolumn{1}{l}{PWA+QE} & \multicolumn{1}{c}{4096} & \multicolumn{1}{c}{\textbf{81.7}} & \multicolumn{1}{c}{\textbf{89.2}} & \multicolumn{1}{c}{\textbf{80.6}} & \multicolumn{1}{c}{\textbf{84.7}} \\
    \toprule[1.5pt]
    \end{tabular}%
  \label{pre-trained}%
\end{table*}%

\subsubsection{Dimensionality reduction}
In order to get shorter representations, we compress the $N\times C$-dimensional  aggregated representation $\psi(I)$ by PCA and whitening process.
Table~\ref{dim} reports the performance of representations with varying dimensionality, M=128 to 4096.
We do not reduce the final representation into higher dimensionality because of  the limited number of images in Oxford5k and Paris6k datasets.
We select N=25 part detectors to aggregate the convolutional features in this experiment.

The results show that the performance boosts gradually with the increase of dimensionality and  the best performance is achieved at 4096 dimensionality.
%The discriminative information of representation is damaged by the compression.
We get the consistent conclusion with other methods,  the compression leads to the loss of discriminative information and performance degradation.
The previous works~\cite{spoc,rmac,crow} aggregate convolutional features as compressed representations with dimensionality under 512, but our PWA representation has more choice of dimensionality.
 %and the compressed representations with lower dimensionality lose more discriminative information.
Compared with~\cite{spoc,rmac,crow}, our PWA methods can generate representations with both low and high dimensionality and achieve better performance on most datasets.
The dimensionality of PWA representation can be chosen according to the tradeoff between performance and efficiency  on different tasks.

\subsection{Comparison with the state-of-the-art}
\begin{table*}[htbp]
  \centering
  \caption{Accuracy comparison with the state-of-the-art supervised methods. Employing the convolutional layer features of fine-tuned network~\cite{fine_tune_3}, we achieve the comparable performance with the state-of-the-art methods with end-to-end supervised training.}
    \vspace{0.5em}
    \begin{tabular}{rrrrrr}
     \toprule[1.25pt]
          &       & \multicolumn{4}{c}{\textbf{Datasets}} \bigstrut\\
\cline{3-6}    \multicolumn{1}{l}{\textbf{Method}} & \multicolumn{1}{c}{\textbf{Dimensionality}} & \multicolumn{1}{c}{Oxford5k} & \multicolumn{1}{c}{Paris6k} & \multicolumn{1}{c}{Oxford105k} & \multicolumn{1}{c}{Paris106k} \bigstrut[t]\\
    \toprule[1pt]
    \multicolumn{1}{l}{NetVLAD~\cite{netvlad}} & \multicolumn{1}{c}{4096} & \multicolumn{1}{c}{71.6} & \multicolumn{1}{c}{79.7} &\multicolumn{1}{c}{---} & \multicolumn{1}{c}{---}  \\
    \multicolumn{1}{l}{CNNBoW~\cite{fine_tune_1}} & \multicolumn{1}{c}{512} & \multicolumn{1}{c}{79.7} & \multicolumn{1}{c}{83.8} &\multicolumn{1}{c}{73.9} & \multicolumn{1}{c}{76.4}  \\

    \multicolumn{1}{l}{DeepRepresentation~\cite{fine_tune_3}} & \multicolumn{1}{c}{2048} & \multicolumn{1}{c}{86.1} & \multicolumn{1}{c}{94.5} & \multicolumn{1}{c}{82.8} & \multicolumn{1}{c}{90.6}  \bigstrut[b]\\
    \hline
    \multicolumn{1}{l}{Previous state-of-the-art} & \multicolumn{1}{l}{} & \multicolumn{1}{c}{86.1} & \multicolumn{1}{c}{94.5} &\multicolumn{1}{c}{82.8} & \multicolumn{1}{c}{90.6}  \bigstrut\\
    \hline
    \multicolumn{1}{l}{PWA} & \multicolumn{1}{c}{2048} & \multicolumn{1}{c}{\textbf{87.8}} & \multicolumn{1}{c}{\textbf{94.9}} &\multicolumn{1}{c}{\textbf{82.8}} & \multicolumn{1}{c}{\textbf{91.0}} \bigstrut[t]\\
    \toprule[1.25pt]
    \end{tabular}%
  \label{fine-tuned}%
\end{table*}%

\subsubsection{Unsupervised methods}
In the first part of Table~\ref{pre-trained}, we compare our PWA  method using pre-trained VGG16~\cite{VGG} with the state-of-the-art unsupervised methods, which employ global representations of images.  Our PWA representation significantly outperform them on all four standard retrieval datasets.
In particular, the gain is more than 8.3\% in mAP on Oxford5k and Oxford105k datasets.
The results  demonstrate that our PWA representation weighted by the selected ``probabilistic proposals'' is effective  and  discriminative for image retrieval.
Our 512-dimensional PWA representation is comparable with the previous state-of-the-art, and its results are only lower than R-MAC~\cite{rmac} on Paris6k.
The PWA representation with higher dimensionality (such as 1024, 2048 and 4096)  consistently outperform all of them on all datasets.

We compare other methods that contain query  expansion (QE) and spatial verification stages with our approach in the second part of Table~\ref{pre-trained}.
In the experiments, we use average query expansion (QE)~\cite{qe}  computed by the top 10 query results.
%Query expansion improves the performance at low extra cost.
Our PWA+QE method performs better than the related works~\cite{rmac,crow} on all datasets.
Although the approximate max pooling localization  (AML) process in R-MAC~\cite{rmac} requires a costly verification stage and the extra memory storage, our PWA+QE still achieves better performance than R-MAC+AML+QE.
%Despite not requiring a costly spatial verification stage, our method is on equal foot or even improve the state-of-the-art on most datasets.
%The previous methods~\cite{qe_3mikulik,qe_6tolias} that perform better than us on Oxford5k and Oxford105k require the time-consuming spatial verification process or perform some learning on the target dataset respectively, while we use a single universal model without costly spatial verification.
DSM~\cite{DSM} uses handcrafted features(SIFT) and achieves better performance by employing additional time-consuming reranking processes ,i.e., spatial verification in tf-idf and k-NN reranking.
%It uses handcrafted features(SIFT) which are helpful in describing small objects.

%\begin{table*}[htbp]
%  \centering
%  \caption{Accuracy comparison with the  work~\cite{fine_tune_3} without shifting and fully connected layers. Our PWA method significantly improves the performance.}
%    \vspace{0.5em}
%    \begin{tabular}{rrrrr}
%     \toprule[1.25pt]
%          &       & \multicolumn{3}{c}{\textbf{Datasets}} \bigstrut\\
%    \cline{3-5}    \multicolumn{1}{l}{\textbf{Method}} & \multicolumn{1}{l}{\textbf{Dimensionality}} & \multicolumn{1}{l}{Oxford5k} & \multicolumn{1}{l}{Paris6k} &
%    \multicolumn{1}{l}{Holidays} \bigstrut[t]\\
%    \toprule[1pt]
%    \multicolumn{1}{l}{Gordo et al.~\cite{fine_tune_3} w/o Shift+FC} & \multicolumn{1}{l}{2048} & \multicolumn{1}{l}{78.7} & \multicolumn{1}{l}{89.7} & \multicolumn{1}{l}{89.1} \bigstrut[b]\\
%    \multicolumn{1}{l}{PWA} & \multicolumn{1}{l}{4096} & \multicolumn{1}{l}{\textbf{84.8}} & \multicolumn{1}{l}{\textbf{92.0}} & \multicolumn{1}{l}{\textbf{89.2}} \bigstrut[t]\\
%    \toprule[1.25pt]
%    \end{tabular}%
%  \label{fine-tuned2}%
%\end{table*}%

\subsubsection{Supervised methods with end-to-end training}
We also compare our  method  with the current state-of-the-art supervised methods  containing end-to-end training process \cite{netvlad,fine_tune_1,fine_tune_3}) in Table~\ref{fine-tuned}.
In order to compare with them, we employ convolutional layers features of fine-tuned ResNet101 from the work~\cite{fine_tune_3}.
Because these methods~\cite{fine_tune_1,fine_tune_3} map the final representation by the supervised methods for similarity evaluation, we also map the PWA representations for comparison purposes.
In order to keep consistently unsupervised, we utilize the unsupervised IME layer~\cite{IME} to map our PWA representations for similarity evaluation.
%While the proposed PWA method does not need the end-to-end supervised training,

The results show that our unsupervised PWA representation outperforms the state-of-the-art supervised methods~\cite{netvlad,fine_tune_1,fine_tune_3} on all datasets.
%In this experiment, we compare the proposed method with the result of  work~\cite{fine_tune_3} that does not contain multi-resolution process which bring extra computational cost at feature extract time (approximately three times cost for three resolutions).
%The scores of PWA are only lower than that of the work~\cite{fine_tune_3} on Paris6k and Holidays.
%In the work~\cite{fine_tune_3}, the PCA is replaced by the shifting and fully connected layers.
%The optimal weights for them are learned end-to-end.
%Compared with it, the strategy of dimensionality reduction and whitening  in our method is unsupervised.
%In Table~\ref{fine-tuned2}, we compare our PWA method with the work~\cite{fine_tune_3} which not containing shifting and fully connected layers.
%As shown in Table~\ref{fine-tuned2}, our PWA method performs better than the network without shifting and fully connected layers~\cite{fine_tune_3}.
%The results demonstrate that the supervised shifting and fully connected layers are important for the performance of this fine-tuned network, while we only use the convolutional layers of it.
%We will explore the more effective method to reduce the dimensionality of representation in the future work.
Furthermore, the effectiveness of the supervised methods  is heavily relied on the collected training set.
%The images in Oxford5k and Paris6k datasets are almost landmarks. However most images in Holidays dataset are photographs of landscapes.
%The networks~\cite{fine_tune_1,fine_tune_2,fine_tune_3} that are fine-tuned by the landmark buildings dataset  significantly improve the performance on Oxford5k and Paris6k datasets but only help little on the performance on  Holidays dataset.
However, our unsupervised PWA method can make better use of the convolutional features extracted from both pre-trained and fine-tuned CNN model to represent the images and does not need the further supervised re-training.
Considering the fact that the annotated training dataset is difficult to collect, it is impractical to fine-tune the model for each discrepant task respectively.
Our unsupervised PWA method is very suitable for this condition.
%We selects part detectors by unsupervised strategy and reduce the dimensionality of PWA representation by unsupervised method (PCA).
Our PWA method retains more discriminative  information  of the retrieval object parts and significantly suppress the noise of background, and  better utilizes the convolutional features extracted from both pre-trained and fine-tuned CNN models.
%PWA不仅适用于pre-trained，也适用于finetuned。
%不fintune的网络提升更大，提供了一条新的方向使用pre-train 的cnn，特别是对训练数据不好收集的情况. PWA充分发挥pre-trained 特征的鉴别性。

\section{Conclusion}
In this paper, we propose a novel PWA method for image retrieval.
The key  characteristic of our method is that it employs discriminative part detectors selected by unsupervised strategy to generate ``probabilistic proposals''.
Based on the selected ``probabilistic proposals''  corresponding to special semantic content implicitly, we weight and aggregate the  deep convolutional features  extracted from pre-trained or fine-tuned CNN models.
Due to the selected ``probabilistic proposals''  corresponding to fixed semantic content but not fixed position, we concatenate the regional representations as global PWA representation.
The results show that our PWA representation suppress the noise of background and highlight the discriminative parts and patterns of retrieval objects.
%Our approach outperform the state-of-the-art aggregation methods in image retrieval.
%Experiments on five standard retrieval datasets demonstrate that our unsupervised approach yields superior results to state-of-the-art aggregation methods in image retrieval.

Experiments on four standard retrieval datasets demonstrate that our unsupervised approach outperforms the previous state-of-the-art unsupervised and supervised aggregation methods.
It is worth noting that our  unsupervised PWA method is very suitable and effective for the situation where the annotated training dataset is difficult to collect.

%In our future work, we plan to cluster the initial part detectors to select some more discriminative part detectors and reduce the number of selected part detectors.
%In addition, the end-to-end dimensionality reduction strategy can be considered to improve the performance.

%{%\scriptsize
%\bibliographystyle{aaai}
%\bibliography{mybib}
%}

\section{Acknowledgments}
This work was supported by the National Natural Science Foundation of China under Grant 61531019, Grant 61601462, and Grant 71621002.

\begin{small}
\bibliographystyle{aaai}
\bibliography{mybib}
\end{small}

\end{document}